%% file: main.tex
\documentclass[11pt]{article}
\usepackage[a4paper,margin=2.5cm]{geometry}
\usepackage[T1]{fontenc}
\usepackage[utf8]{inputenc}
\usepackage{mathptmx}
\usepackage{microtype}
\usepackage{amsmath,amssymb}
\usepackage{booktabs}
\usepackage{tabularx}
\usepackage{array}
\usepackage{graphicx}
\usepackage{xcolor}
\usepackage{algorithm}
\usepackage{algpseudocode}
\usepackage[hidelinks]{hyperref}
\usepackage{url}
\usepackage{enumitem}
\setlist{nosep,leftmargin=1.4em}

\newif\ifanon
\anonfalse
\IfFileExists{anon.flag}{\anontrue}{}

\input{generated/numbers.tex}



\newcommand{\Hsel}{H_{\mathrm{sel}}}
\newcommand{\Hval}{H_{\mathrm{val}}}

\title{The Harness as the Only Mutable Surface:\\
Compliance-Bounded Self-Evolution of LLM Agents\\
in Credit Pipelines, with a Measured Admission Gate}

\ifanon
\author{Anonymous author(s)}
\else
\author{Ravil Akhtyamov\\[2pt]
\small Digital Economy Lab\\
\small \texttt{latam@digitaleconomylab.org}\quad ORCID 0000-0002-0783-8573}
\fi
\date{}

\begin{document}
\maketitle

\begin{abstract}
\noindent
Self-improving large language model (LLM) agents can adapt a credit pipeline to a changed rule, but an agent that rewrites itself destroys the artefact a supervisor reviews: a named change, a recorded test, an approval. We argue that self-evolution is reviewable only if it is confined to the runtime harness---instruction text, tool-call logic and primitive composition---while model weights stay fixed, so that every adaptation is a diff with a cause and a test attached. We give a dual-loop engine built on that bound, in which a local loop edits one primitive and a global loop replaces primitives from a typed library, and both submit to one admission gate that writes a hash-chained record before deployment. We then measure the gate in simulation, with a simulated agent and a seeded-search proposer rather than language models. Across three families of supervisory re-interpretation at three severities, \nseeds{} seeds each, the gated dual loop admitted \gadmdual{} of \gpropdual{} candidate changes, none of which worsened error on held-out history, and restored the false-positive rate to the oracle level without raising missed flags in all low- and mid-severity cells. With the gate replaced by the check an unbounded system applies---fewer errors visible in recent traces---the same loops admitted \gfanogate{} harmful changes and left missed flags above 10\% in \gmtennogate{} of \ncells{} runs: false positives fell because the screen was loosened. Evaluated on pre-shift labels, the gate rejected every candidate, so a re-interpretation must be encoded as a rule that relabels history. Parametric and scope shifts were repaired locally, a structural one only by primitive replacement; at the highest structural severity the gate's fixed tolerance blocked the correct replacement in half the seeds. We map the mechanisms to the EU AI Act's provisions for high-risk credit scoring and note that the April~2026 US model-risk guidance excludes agentic AI from its scope.
\end{abstract}

\noindent\textbf{Keywords:} LLM agents; self-evolving agents; credit risk; model risk management; EU AI Act; auditability; RegTech; benchmark

\section{Introduction}

Credit underwriting is moving from monolithic scoring models to pipelines of specialised LLM agents that query registries, screen counterparties and draft decisions. When a statute or a supervisory interpretation changes, such a pipeline must change too. Work on self-evolving agents promises to automate that adaptation~\cite{fang2025survey,gao2025survey}, but it was built to raise a task metric on open benchmarks, not to leave behind what a supervisor reviews.

The difficulty is not that self-evolution is unsafe in general. It is that model-risk governance presumes a versioned object with an owner, a test result and an approval~\cite{sr262,euaiact}, and continuous self-modification dissolves all three. The question is therefore not whether to bound self-evolution but which bound preserves the reviewable artefact.

Our answer is that the runtime harness is the only legitimate mutable surface, \emph{because} it is the auditable one. Weights cannot be diffed into named changes; a harness can. If the engine may rewrite instruction text, tool-call logic and the composition of typed primitives, and nothing else, then ``the agent changed'' stops being an event that requires re-validating a model and becomes a text diff with a cause and a test attached. The self-evolution literature treats the harness as one surface among several to optimise over~\cite{selfharness,harnessx,gepa}; we treat it as the only one an institution may let evolve.

A bound is only as good as the control that enforces it. In an earlier presentation of this design the admission gate---the mechanism that distinguishes it from unbounded methods---was specified but not measured, and three referees said so. This paper is the measurement. We claim:

\begin{enumerate}
\item \textbf{A position and an architecture} (Sections~\ref{sec:harness}--\ref{sec:engine}). Harness-only evolution with fixed, digest-pinned weights (a design commitment: no model is pinned in our simulation); a local and a global loop sharing one retrospective pool and one gate; an admission rule with a value for every threshold (Table~\ref{tab:constants}); a hash-chained log written before deployment. We separate what the log makes auditable (the configuration change) from what it does not make reproducible (downstream behaviour of a hosted model).
\item \textbf{A measured gate} (Sections~\ref{sec:design}--\ref{sec:results}). A regulatory-shift track with three families and three severities of re-interpretation, \nseeds{} seeds per cell and ten arms, including a no-gate arm, a gate on stale labels, oracle and manual updates and a compute-matched local-only arm. We report missed-flag rates on every row, candidate counts, false admissions audited on held-out data, one worked rejection, and the gate's own failure mode.
\item \textbf{An answer to where the labels come from} (Section~\ref{sec:labels}). After a re-interpretation the retrospective pool must be relabelled by the new rule, or a correct adaptation is indistinguishable from a regression. This makes the adaptation supervised at the level of the rule and autonomous only at the level of the harness, and we position the autonomy claim accordingly.
\item \textbf{A regulatory mapping} (Section~\ref{sec:reg}) onto the EU AI Act's provisions on predetermined changes, logging, human oversight, robustness and change management, and an account of the gap the April~2026 US guidance leaves for agentic systems.
\item \textbf{A released benchmark} (Section~\ref{sec:ref}), CACE-Bench, whose reference run on decision availability we now report with missed-flag rates and \refnseeds{} seeds.
\end{enumerate}

What we do \emph{not} claim is equally specific. The proposer in our experiments is a seeded search over a stated edit space, not a language model, and the typed library contains the correct replacement primitive by construction. We therefore measure whether the bound and the gate select correct changes and reject harmful ones, not whether an LLM can invent the changes. All data are synthetic. No supervisor has reviewed the mapping.

\section{Related work}
\label{sec:related}

\paragraph{Self-evolving agents.} Agents that improve themselves now span memory~\cite{reflexion}, prompts~\cite{promptbreeder,gepa}, harnesses~\cite{selfharness}, typed harness primitives co-evolved with model parameters~\cite{harnessx}, workflows~\cite{evoagentx}, and code or architecture~\cite{adas,godel,dgm}; two surveys organise the field by what evolves and where the loop closes~\cite{fang2025survey,gao2025survey}. Our local loop follows the weakness-mining, proposal and validation cycle of Self-Harness~\cite{selfharness}; our global loop follows the primitive replacement of HarnessX~\cite{harnessx}, without HarnessX's co-evolution of model parameters, which is precisely what our bound excludes. We do not claim either loop as new. Table~\ref{tab:positioning} places the work on the axes a regulated deployment cares about. The field closes its loops on a task metric and produces no change record intended for a third party; that is a governance gap, not a capability gap, and we make no claim that our engine optimises better than any of these systems.

\begin{table}[t]
\centering\small
\caption{Positioning against representative self-evolution systems. \emph{Gate}: is a change admitted only after an explicit, recorded test? \emph{Record}: is a change artefact produced for an external reviewer? \emph{Gate measured}: are admission and rejection statistics reported? $\sim$: a validation step exists but is neither recorded for review nor reported as a control.}
\label{tab:positioning}
\begin{tabular}{lllll}
\toprule
System & What evolves & Gate & Record & Gate measured\\
\midrule
Reflexion~\cite{reflexion} & memory & -- & -- & -- \\
PromptBreeder~\cite{promptbreeder} & prompts & -- & -- & -- \\
GEPA~\cite{gepa} & prompts & $\sim$ & -- & -- \\
Self-Harness~\cite{selfharness} & harness & $\sim$ & -- & -- \\
HarnessX~\cite{harnessx} & primitives, weights & $\sim$ & -- & -- \\
EvoAgentX~\cite{evoagentx} & workflows & $\sim$ & -- & -- \\
ADAS~\cite{adas}, DGM~\cite{dgm}, G\"odel Agent~\cite{godel} & code / architecture & -- & -- & -- \\
This work & harness only & yes & yes, hash-chained & yes \\
\bottomrule
\end{tabular}
\end{table}

\paragraph{Evaluation of agents in finance.} We build on a three-level quality-metric system for LLM agents in the credit pipeline introduced in prior work~\cite{akhtyamov2026metrics}; it is not a contribution of this paper. Every number reported here is computed against deterministic labels; no LLM judge is involved anywhere in the results.

\paragraph{Model-risk governance.} US supervisory model-risk guidance was revised on 17~April~2026 (SR~26-2; OCC Bulletin 2026-13; FDIC FIL-15-2026), rescinding SR~11-7 and stating that generative and agentic AI models are outside its scope~\cite{sr262,occ202613}. The EU AI Act classifies creditworthiness assessment of natural persons as high-risk~\cite[Annex~III, point~5(b)]{euaiact} and regulates change in systems that continue to learn~\cite[Art.~43(4)]{euaiact}. Section~\ref{sec:reg} develops both.

\section{Problem, threat model and requirements}
\label{sec:problem}

We consider a multi-agent credit pipeline---client assistant, underwriter, compliance and legal agents---whose agents execute \emph{typed primitives}: named, individually testable units of behaviour such as registration verification or the beneficial-owner check. A primitive is defined by its instruction text, its tool-call contract and its output schema. The set of primitives available to an agent, with their instruction texts and parameters, is that agent's \emph{runtime harness}. It is the only object the engine may modify and the granularity at which every audit record is written. The benchmark is jurisdiction-parametrised; its reference run (Section~\ref{sec:ref}) covers seven Latin American jurisdictions, and nothing in the engine depends on a particular national statute.

\paragraph{Threat model.} We design against five failure modes.
\textbf{F1 Silent drift}: a change improves the target metric while degrading an unmeasured one; mitigated by the non-regression clause, and only to the extent that the degraded dimension is measured.
\textbf{F2 Retro-test overfitting}: a candidate is tuned until it passes the pool; mitigated by a pool split the gate never sees ($\Hval$) and a held-out post-shift window.
\textbf{F3 Untraceable change}: mitigated by making the log write a precondition of deployment.
\textbf{F4 Objective capture}: evolution objective and audit evidence from the same judge; mitigated by computing both on deterministic labels only.
\textbf{F5 Exogenous model change}: the provider updates a hosted model beneath a fixed harness; \emph{not} mitigated by our design (Section~\ref{sec:harness}).

\paragraph{Requirements.} (R1) \emph{Measurability}: every step yields signals localisable to a primitive. (R2) \emph{Auditability}: each change carries a cause, a test result and a deployment record in an append-only log. (R3) \emph{Bounded autonomy}: adaptation must not degrade previously correct behaviour beyond a stated tolerance, and weights must not change. (R4) \emph{Regulatory correspondence}: controls must correspond to identifiable obligations (Section~\ref{sec:reg}).

\section{The harness as the governance primitive}
\label{sec:harness}

The engine may rewrite instruction text, tool-call logic and primitive composition; it may not fine-tune, and the served model is pinned by digest. This preserves the object a supervisor knows how to review. A change under model-risk governance is expected to have a cause, a test, an approver and a rollback path. A weight update supplies none of these naturally: it is continuous, not decomposable into named changes, and cannot be diffed. A harness edit supplies all of them by construction: it is a diff against a named primitive, reproducible from the log, and revertible by writing the previous version back.

Two properties must be kept apart, because a referee rightly observed that we had conflated them. \emph{Auditability of the change} is delivered: every harness version is a canonical serialisation with a SHA-256 digest, every gate decision names the digests before and after, and the log is hash-chained so that removing or reordering a record breaks verification (all \nlogsok{} of the \nlogs{} logs written in our experiments, \nrecords{} records in all, verify). The record names a deploying identity; in the simulation it is the gate's service identity, and a human approval step before deployment is a deployment choice we do not simulate. \emph{Reproducibility of behaviour} is not delivered by the same mechanism: identical harnesses on a hosted model can behave differently if the provider changes the model, the decoding settings drift, a tool returns different data, or retrieval state moves. Weight attestation binds our changes and not the provider's (F5). For a deterministic pipeline---as in our benchmark---the two coincide; for a hosted LLM, behaviour reproducibility additionally requires pinned model versions and decoding parameters, recorded tool responses, and canary probes on a fixed prompt set.

The cost of the bound is real: the system cannot learn anything that instruction text and primitive composition cannot express. A task the base model cannot do falls outside the loop and remains engineering work. We regard that as the right trade for regulated deployment, not as a free one.

\section{The engine}
\label{sec:engine}

\subsection{Admission rule}

Let $h$ be the deployed harness, $h'$ a candidate, $\Hsel$ the half of the retrospective pool the gate may use, labelled under the \emph{current} rule (Section~\ref{sec:labels}), $C \subseteq \Hsel$ the cases $h$ answers correctly, and $k \in \{\mathrm{FP},\mathrm{miss}\}$ the error kind of the cluster that caused the proposal. Over the cases in $\Hsel$ on which an error of kind $k$ is possible (true \textsc{clear} for FP, true \textsc{flag} for misses), let $b$ count cases where $h$ errs and $h'$ does not, $c$ the reverse, and $n_k$ their number. The paired reduction is $\hat g = (b-c)/n_k$ with variance estimate $\widehat{\mathrm{var}} = (b+c)/n_k^2 - (b-c)^2/n_k^3$. The candidate is admitted iff
\begin{equation}
\label{eq:gate}
\underbrace{\hat g - z_{0.975}\sqrt{\widehat{\mathrm{var}}} > 0}_{\text{significant improvement}}
\quad\wedge\quad
\underbrace{\frac{|\{x \in C : h'(x) \neq y(x)\}|}{|C|} \le \epsilon}_{\text{non-regression on } C}.
\end{equation}
The improvement margin is statistical rather than a fixed $\delta$, so 0.1 and 5 percentage points are no longer alike to the gate. The tolerance $\epsilon = 0.005$ is not zero because execution noise makes strict non-regression unsatisfiable: any change that alters the rule's output on a case can meet that case's noise draw and turn a correct answer wrong. Section~\ref{sec:eps} measures what this choice costs.

Admission is a write, not only a decision. Before a candidate reaches the runtime the gate appends $\langle$sequence, stream position, loop, cause, diff, digest of $h$, digest of $h'$, rendered text of $h'$, $b$, $c$, $n_k$, $\hat g$ and its interval, $|C|$, regressions on $C$, decision, deploying identity, hash of the previous record$\rangle$ and the record's own hash. Rejections are written with the same schema.

\subsection{Loops}

Operational feedback arrives in windows of 500 cases. Every flagged case is reviewed by an analyst, so its false positive becomes visible; cleared cases are audited at 5\%, so most misses stay invisible. This asymmetry is deliberate and realistic, and it is what makes an ungated loop dangerous: a proposer guided by visible failures sees false positives and almost no misses.

\begin{algorithm}[t]
\caption{Local loop, per window}
\label{alg:local}
\begin{algorithmic}[1]
\Require window $W$, deployed harness $h$, gate pool $\Hsel$, budget $B_\ell=3$, clusters $K=3$, escalation threshold $N=3$
\State $F \gets$ visible errors in $W$ clustered by (hit type, error kind); keep the $K$ most frequent
\For{each cluster $f \in F$}
  \State $P \gets$ one-step edits to the deployed primitive (Table~\ref{tab:constants}), shuffled; keep $B_\ell$
  \For{each $h' \in P$}
    \State write record; \textbf{if} $h'$ satisfies Eq.~\eqref{eq:gate} on $\Hsel$ \textbf{then} $h \gets h'$; \textbf{break}
  \EndFor
  \State $\mathit{stall}[f] \gets 0$ if admitted, else $\mathit{stall}[f]+1$
  \If{$\mathit{stall}[f] \ge N$} \textsc{Escalate}$(f)$ to Algorithm~\ref{alg:global}; $\mathit{stall}[f] \gets 0$ \EndIf
\EndFor
\end{algorithmic}
\end{algorithm}

\begin{algorithm}[t]
\caption{Global loop, on escalation (dual loop) or on threshold breach (global-only arm)}
\label{alg:global}
\begin{algorithmic}[1]
\Require cluster $f$, deployed harness $h$, typed library $\mathcal{L}$, budget $B_g=6$
\State $P \gets \{(p,\theta): p \in \mathcal{L}\setminus\{p_h\},\ \theta \in \Theta\}$ with $h$'s scope, shuffled; keep $B_g$
\For{each $h' \in P$}
  \State write record; \textbf{if} $h'$ satisfies Eq.~\eqref{eq:gate} on $\Hsel$ \textbf{then} $h \gets h'$; \textbf{return}
\EndFor
\end{algorithmic}
\end{algorithm}

The \emph{local loop} (Algorithm~\ref{alg:local}) may change one parameter or one clause of the deployed primitive per candidate. The \emph{global loop} (Algorithm~\ref{alg:global}) may replace the primitive with another from a typed library---here, a check on the stake held through the first-layer vehicle versus a check on the effective stake along the ownership chain---so a replacement is still a discrete, nameable change. It has no privileged path to deployment: the same gate applies. In the global-only arm the loop is triggered when the visible false-positive share of reviewed flags exceeds its pre-shift level by $\tau = 5$ percentage points for two consecutive windows. Table~\ref{tab:constants} lists every constant.

\begin{table}[t]
\centering\small
\caption{Every free parameter of the engine and the experiment, with the value used. All are constants in \texttt{evolve.py}.}
\label{tab:constants}
\begin{tabularx}{\linewidth}{lX}
\toprule
Parameter & Value\\
\midrule
Stream & 23{,}000 screening alerts; shift at $T = 70\%$; adaptation on $[T, 85\%)$; held-out evaluation on $[85\%, 100\%]$\\
Retrospective pool & pre-$T$ stream; even-indexed cases $\Hsel$ (gate), odd-indexed $\Hval$ (audit only)\\
Window / cycle & 500 cases; flags reviewed 100\%; clears audited 5\%\\
Execution noise & outcome flipped with probability 0.02 per case, draw shared across harness versions\\
Local loop & $B_\ell = 3$ candidates per cluster, $K = 3$ clusters per cycle; for a false-positive cluster: remove its hit type from scope, or $\theta + \{0.05, 0.10, 0.15, 0.20, 0.25\}$; for a miss cluster: add its hit type, or $\theta - \{0.05, 0.10, 0.15\}$\\
Escalation & $N = 3$ consecutive cycles without admission for a cluster\\
Global loop & $B_g = 6$ candidates; library $\{$first-layer stake, effective stake$\}$; $\Theta = \{0.15, 0.20, 0.25, 0.30, 0.35, 0.40, 0.50\}$\\
Gate & paired 95\% lower bound of reduction $>0$; $\epsilon = 0.005$; sensitivity at $\epsilon = 0.010$ (structural family)\\
Global-only trigger & visible flag-FP share above pre-$T$ level by $\tau = 0.05$ for 2 windows\\
Manual update & correct harness deployed 2{,}000 cases after $T$ (an assumption, not a measurement)\\
Agent & a deterministic rule applying the harness, plus the execution noise above; no language model\\
Proposer & seeded random choice from the edit space above; no language model\\
Seeds & \nseeds{} per cell; 3 families $\times$ 3 severities $=$ 9 cells, \ncells{} runs\\
\bottomrule
\end{tabularx}
\end{table}

\section{The regulatory-shift track}
\label{sec:design}

\subsection{World and rules}

The population is the screening-alert queue: each case carries a name hit of one of five types (formal designation, high- and low-grade adverse media, foreign and domestic politically exposed person) and a beneficial-ownership chain of depth one to three. The compliance primitive returns \textsc{flag} or \textsc{clear}. Ground truth is a versioned rule. Before $T$ (rule v1): flag when the hit type is in scope and the stake held through the first-layer vehicle is at least 25\%. From $T$ (rule v2) one of three re-interpretations applies:
\begin{itemize}
\item \textbf{Thr} (parametric): the ownership threshold rises to 30, 40 or 50\%.
\item \textbf{Scope} (clause): low-grade adverse media; then all adverse media; then also domestic PEP status no longer qualify on their own.
\item \textbf{Struct} (structural): the threshold applies to the \emph{effective} stake, the product of stakes along the chain, with multi-layer chains in 30, 50 or 70\% of cases. No threshold or scope edit to the first-layer check can express this rule.
\end{itemize}
All three make the v1 harness over-flag, so the observable symptom is the same---a rise in false positives---while the correct repair differs in kind.

\subsection{Where the labels come from after a re-interpretation}
\label{sec:labels}

A referee observed that, as we had described it, the retrospective test was circular: the pre-shift pool carries labels under the interpretation the shift replaced, so a correct adaptation must look like a regression on it. The objection is correct, and the answer is a design commitment rather than a detail. At $T$ the pool is \textbf{relabelled by applying rule v2 to the stored case facts}. In a deployment this is the compliance function encoding the supervisor's re-interpretation as an executable rule---nobody relabels cases by hand, but a person authors the rule. The engine therefore searches for a harness edit autonomously while the target of the search is set by people: the adaptation is \emph{supervised at the level of the rule and autonomous at the level of the harness}. We position the autonomy claim accordingly. The ``gate on stale labels'' arm, which keeps the v1 labels, shows empirically what happens otherwise (Section~\ref{sec:stale}).

The relabelling assumes the rule encodes the re-interpretation correctly; a mis-encoded rule would be faithfully optimised towards. That is the right place for the risk to sit---the rule is a short, reviewable artefact authored by an accountable function---but it is a risk.

\subsection{Arms}

All arms see the same cases, the same noise draws and the same feedback model; what each observes depends on its own harness. \emph{Healthy}: no shift, v1 harness. \emph{Degraded}: shift, no adaptation. \emph{Oracle}: the correct v2 harness from $T$. \emph{Manual}: the correct harness after a 2{,}000-case lag, standing in for an engineer who rewrites the prompt after an alert. \emph{No gate}: the dual loop with Eq.~\eqref{eq:gate} replaced by the check an unbounded self-evolution system applies---admit if the candidate makes fewer errors visible in the current window. \emph{Gate on stale labels}: the dual loop with $\Hsel$ under v1 labels. \emph{Global only}, \emph{local only}, and \emph{local only, compute-matched}, whose per-cycle budget is multiplied, per run, by the ratio of the dual loop's proposals to the local-only arm's (rounded up; between \cmScaleMin{} and \cmScaleMax{} in our runs). \emph{Dual loop}: the full engine.

\subsection{Metrics}

On the held-out window, with \textsc{flag} as the positive class: false-positive rate $=\mathrm{FP}/(\mathrm{FP}+\mathrm{TN})$; missed-flag rate $=\mathrm{FN}/(\mathrm{FN}+\mathrm{TP})$ and recall $=1-$missed-flag rate; \emph{regression rate} $=$ the share of held-out cases answered correctly by the degraded harness that the final harness answers wrongly (a paired, case-level measure). During adaptation: \emph{excess errors} $=$ true errors in $[T,85\%)$ minus those of the oracle arm; \emph{recovery window} $=$ the first window from which the arm's error rate stays within one percentage point of the oracle's (7 means never). For the gate: candidates proposed, admitted and rejected by reason; \emph{false admissions}, admitted changes that increased true error on $\Hval$, which the gate never sees; and \emph{oracle rejections}, rejections of the exactly correct harness. Means are over seeds with 95\% $t$-intervals, clipped to $[0,100]$. The held-out window holds 3{,}450 alerts per run; its numbers of true \textsc{clear} and true \textsc{flag} cases under the post-shift rule, which are the denominators of the two rates, are given per cell in Table~\ref{tab:grid}.

\section{Results}
\label{sec:results}

Table~\ref{tab:main} reports the mid-severity cell of each family for all ten arms; Table~\ref{tab:grid} the whole grid for six arms; Table~\ref{tab:gate} the gate statistics over all \ncells{} runs.

\begin{table}[p]
\centering\scriptsize
\caption{Mid severity, held-out window, mean over seeds with 95\% $t$-intervals. FP and missed-flag rates in \%; intervals clipped to $[0,100]$. For the no-gate arm the missed-flag interval is replaced by the range \{min--max\} across seeds because its distribution is bimodal. Regression on held-out cases previously answered correctly, in \%. Gate: candidates proposed / admitted / false admissions audited on $\Hval$, summed over seeds.}
\label{tab:main}
\setlength{\tabcolsep}{4pt}
\begin{tabular}{lllcccc}
\toprule
Arm & FP & Missed flag & Regr. & Excess err. & Recov. win. & Gate: prop. / adm. / false\\
\midrule
\input{generated/tab_main.tex} 
\bottomrule
\end{tabular}
\end{table}

\begin{table}[t]
\centering\footnotesize
\caption{Full grid, held-out window: FP / missed-flag rate in \%, mean over \nseeds{} seeds. $n$: true \textsc{clear} / true \textsc{flag} cases per seed under the post-shift rule, the denominators of the two rates.}
\label{tab:grid}
\begin{tabular}{lllcccccc}
\toprule
Family & Severity & $n$ clear / flag & Degraded & No gate & Global only & Local only & Dual & Oracle\\
\midrule
\input{generated/tab_grid.tex} 
\bottomrule
\end{tabular}
\end{table}

\begin{table}[t]
\centering\small
\caption{Gate statistics over all \ncells{} runs (9 cells $\times$ \nseeds{} seeds). Rejections are counted by the first clause of Eq.~\eqref{eq:gate} that fails (improvement is checked first): no significant improvement / regression on $C$. $^\dagger$No-gate rejections are by the trace check. False adm.: admitted changes that increased true error on $\Hval$. Oracle rej.: rejections of the exactly correct harness. Miss$>$10\%: runs whose final missed-flag rate exceeds 10\%. Max adm.: largest number of admissions in one run.}
\label{tab:gate}
\footnotesize\setlength{\tabcolsep}{3pt}
\begin{tabular}{lcccccccc}
\toprule
Arm & Prop. & Adm. & Rej.: no gain & Rej.: regr. & False adm. & Oracle rej. & Miss$>$10\% & Max adm.\\
\midrule
\input{generated/tab_gate.tex} 
\bottomrule
\end{tabular}
\end{table}

\subsection{The gate is what keeps the screen from being loosened}

Every arm that recovers brings the false-positive rate down to the oracle's. What separates them is the missed-flag rate, which the earlier presentation did not report. With the gate, the dual loop ends at the oracle's missed-flag rate in all three families (Table~\ref{tab:main}; \missthrdual\%, \missscopedual\% and \missstructdual\% against the oracle's \missthroracle\%, \missscopeoracle\% and \missstructoracle\%), with \gfadual{} false admissions over \gadmdual{} admissions in \ncells{} runs (Table~\ref{tab:gate}). Without it, the same proposers, fed the same windows, admitted \gfanogate{} changes that increased true error on $\Hval$; the final missed-flag rate exceeded 10\% in \gmtennogate{} of \ncells{} runs, reaching a mean of \missscopenogate\% in the mid scope cell while its false-positive rate matched the oracle's. This is the failure referees anticipated: under scope and structural re-interpretations, false positives fall because flagging stops. Under the threshold family the visible-failure check happens to point in the right direction---raising the threshold is the repair---and the no-gate arm's missed-flag rate stays near the oracle's. It is not hypothetical here; it is what a loop closed on visible failures does when misses are mostly invisible, and Eq.~\eqref{eq:gate}, evaluated on fully labelled history, is what stops it. The no-gate arm also churns: up to \gmaxadmnogate{} admissions in a run against at most \gmaxadmdual{} for the dual loop, and more excess errors during adaptation than leaving the pipeline degraded (\exscopenogate{} against \exscopedegraded{} in the mid scope cell).

\subsection{Stale labels make the gate refuse every correct change}
\label{sec:stale}

The gate on v1 labels rejected all \gpropstalegate{} of its candidates, every one for lack of significant improvement---including the exactly correct harness \gorjstalegate{} times; every one of the \staleBothFail{} also failed the non-regression clause--- and left the pipeline at the degraded false-positive rate in every cell. Under the old interpretation a correct repair does not reduce errors; it creates them. The circularity the referee identified is therefore real, and relabelling the pool by the new rule (Section~\ref{sec:labels}) is a necessary part of the design, not an implementation choice.

\subsection{Division of labour between the loops}

The ablation separates the loops cleanly by the kind of re-interpretation (Table~\ref{tab:grid}). Parametric (\emph{thr}) and clause (\emph{scope}) shifts are repaired by the local loop alone, and the global-only arm, whose library offers only primitive replacements, cannot repair them. The structural shift is the reverse: the local-only arm proposed \famstructlocalonlyprop{} candidates across the three structural cells and admitted \famstructlocalonlyadm{}; \famstructlocalonlyreg{} were rejected for regression, because no threshold on the first-layer stake reproduces a rule on the effective stake; the local-only arm stays at the degraded rate, and the compute-matched local-only arm, with its budget scaled to the dual loop's proposal count, does no better (\fpstructlocalonlycm\% against \fpstructdegraded\% in the mid cell). Repair comes only through primitive replacement. Local-only repair of the scope family is not perfect either: in the mid cell it recovered in \loScopeMidRecovered{} seeds. The dual loop repairs all three families because escalation routes each problem to the loop that can express its fix. This division of labour is, in part, true by construction---the local edit space cannot express a rule on the effective stake, and the library contains one---so what the experiment adds is that escalation finds the right loop without being told the family, and that the gate rejects the inexpressible local candidates rather than admitting the least-bad of them. The dual loop pays for routing in two ways. It recovers more slowly than the right single loop on the structural family, because the local loop must stall for $N$ cycles before escalating (excess errors \excstructmiddual{} against \excstructmidglobalonly{} for global-only in the mid cell). And it makes extra proposals after recovery: residual noise still forms clusters, their candidates are proposed and rejected, and some escalate; none of these was admitted.

The earlier version of this paper reported this division of labour from a single run on different metrics. A referee suggested it was the cleaner finding; the measured version confirms the structure and replaces the anecdote.

\subsection{Recovery speed against a human engineer}

Under our assumed 2{,}000-case lag, the manual update accumulates more excess errors during adaptation than the dual loop in every family at mid severity (Table~\ref{tab:main}), but not everywhere: in the high-severity structural cell, where the dual loop recovers in only half the seeds, it accumulates \excstructhighdual{} against the manual update's \excstructhighmanual. The comparison depends entirely on that lag, which is an assumption; its value is to show where a gated automatic loop helps---time to repair---not to claim the automatic loop is better than a person.

\subsection{Worked rejection}

In the mid structural cell, seed~0, the largest visible error cluster after the shift is false positives on formal designations, and at case \wrT{} the local loop's first candidate is \emph{\wrDiff}: stop treating a designation as a qualifying hit. It is the kind of edit a loop guided only by visible failures finds attractive, since flags it no longer raises can no longer be reviewed as false. On $\Hsel$ under v2 labels it removes \wrImproved{} false positives and introduces \wrWorsened{} among \wrNtarget{} true \textsc{clear} cases, a reduction of \wrGain{} points (95\% interval [\wrGainLo, \wrGainHi]), so the improvement clause passes. But \wrCreg{} of the \wrC{} cases the deployed harness answered correctly (\wrCrate\%) would turn wrong---designated owners above the threshold, now cleared. The record, sequence \wrSeq{} with hash prefix \texttt{\wrHash} chained to the genesis value, is written as \emph{rejected: regression on $C$}. Threshold edits to the first-layer check fail the same clause, because none expresses a rule on the effective stake. After \wrLocalRej{} local rejections the cluster escalates; the global loop's first \wrGlobalRej{} candidates, effective-stake checks at thresholds \wrGlobalRejList, are rejected, and at case \waT{} it proposes \emph{\waDiff}, which reduces false positives by \waGain{} points (lower bound \waGainLo) with \waCreg{} regressions on $C$ (\waCrate\%, attributable to execution noise), and is admitted as record \waSeq. Both records, and every other, are in the released results.

\subsection{What the tolerance costs}
\label{sec:eps}

The gate has a second failure mode, and we report it. In the high-severity structural cell, replacing the primitive changes the rule's output on a large share of cases, and the shared execution noise then turns enough of them wrong that the correct replacement exceeds $\epsilon = 0.005$. The dual loop recovered in \epsrechighA{} seeds, rejecting the exactly correct harness \epsorjhighA{} times, and its mean false-positive rate stayed at \epsfphighA\% (Table~\ref{tab:eps}). Doubling the tolerance to $\epsilon = 0.010$, which we tested on the structural family only, recovers \epsrechighB{} seeds without any false admission. A fixed $\epsilon$ conflates noise with regression; a tolerance scaled to the expected noise on the cases a candidate changes would separate them, and is the next revision of the rule. Until then, $\epsilon$ is a policy parameter: lower values guard more strictly against regression and, as measured here, reject more large correct changes. No harmful change was admitted at either value, so the protective side of the trade-off is not measured by this track.

\begin{table}[t]
\centering\small
\caption{Sensitivity to the non-regression tolerance, structural family, dual loop, held-out window. Recovered: seeds whose final harness is exactly the correct one. Oracle rej.: rejections of the exactly correct harness, summed over seeds.}
\label{tab:eps}
\begin{tabular}{llcccccc}
\toprule
Severity & $\epsilon$ & FP \% & Missed flag \% & Recovered & Oracle rej. & False adm.\\
\midrule
\input{generated/tab_eps.tex} 
\bottomrule
\end{tabular}
\end{table}

\subsection{Harness growth}

A referee asked whether repeated repair accumulates special-case rules. In this track the edit space is parametric, so the harness's length is bounded (its rendered text ranged from \hsizemin{} to \hsizemax{} characters across all runs); what can grow is churn. The gated dual loop made at most \gmaxadmdual{} admissions in a run, the no-gate arm up to \gmaxadmnogate. With an LLM proposer that writes free text, length and rule count become the quantities to bound, and the log already records each version's rendered text for that purpose.

\section{Regulatory correspondence}
\label{sec:reg}

\paragraph{The EU AI Act.} AI systems intended to evaluate the creditworthiness of natural persons or establish their credit score are high-risk~\cite[Annex~III, point~5(b)]{euaiact}; after the Digital Omnibus on AI the corresponding obligations apply from 2~December~2027~\cite{omnibus}. The provision that matters most for self-evolution is Article~43(4): for high-risk systems that continue to learn, changes \emph{pre-determined} by the provider at the initial conformity assessment and described in the technical documentation do not constitute a substantial modification, whereas a change not foreseen in that assessment that affects compliance does~\cite[Arts.~3(23), 43(4)]{euaiact}. A harness-only engine with a fixed edit space, a fixed library and a fixed admission rule is a way of making change \emph{pre-determinable}: the set of reachable harnesses and the rule that admits them can be written into the technical documentation before deployment. Table~\ref{tab:reg} maps the remaining obligations.

\begin{table}[t]
\centering\small
\caption{Obligation, mechanism, artefact, and the status of the evidence in this paper. This is a design argument, not a compliance opinion.}
\label{tab:reg}
\begin{tabularx}{\linewidth}{>{\raggedright}p{2.6cm}>{\raggedright}X>{\raggedright}X>{\raggedright\arraybackslash}p{2.6cm}}
\toprule
Obligation (EU AI Act) & Mechanism & Artefact handed to an examiner & Evidence here\\
\midrule
Predetermined changes, Art.~43(4); substantial modification, Art.~3(23) & Harness-only evolution over a fixed edit space and library & Edit space, library and Eq.~\eqref{eq:gate} in technical documentation & Implemented; bound enforced in code\\
Record-keeping, Art.~12 & Gate writes before deployment; hash chain & Append-only log of admissions and rejections & Implemented; \ncells{} logs verified\\
Accuracy and robustness; feedback loops, Art.~15(1),(4) & Non-regression on $C$; held-out $\Hval$; missed-flag rate tracked & Per-candidate test result & Measured: \gfadual{} false admissions (a weak test; see Section~\ref{sec:threats})\\
Traceability of the model in force & Weights pinned by digest & Model digest per decision & Design only: no model in the simulation\\
Change management in the QMS, Art.~17(1)(a) & Gate as the only path to deployment & Rejected-candidate records & Measured: rejections by reason\\
Human oversight, Art.~14 & Rule authored by the compliance function; escalation to people & Rule version per decision; escalation records & Design only\\
Post-market monitoring, Art.~72 & Windowed metrics on reviewed decisions & Per-window error series & Implemented in simulation\\
Data governance and bias, Art.~10(2)(f)--(g) & None & None & \textbf{Gap}: no protected attributes\\
\bottomrule
\end{tabularx}
\end{table}

\paragraph{The US gap.} The interagency guidance of 17~April~2026 replaced SR~11-7 and OCC Bulletin~2011-12, the framework our earlier presentation leaned on, and states that generative and agentic AI models are not within its scope; the agencies announced a forthcoming request for information on AI use, including generative and agentic AI~\cite{sr262,occ202613}. We read this as strengthening rather than weakening the case for the design. There is, as of writing, no US supervisory template for change control in agentic systems, and the guidance that does exist addresses models whose changes are versioned objects. An architecture that turns agentic adaptation back into versioned, tested, recorded changes is a candidate answer to the question the agencies have left open. We claim no more than that: the guidance does not endorse the design, and our mapping is to the EU text.

Two further limits belong in this section. Bias monitoring under Article~10 has no mechanism in our design and no evidence in our experiments, because the benchmark carries no protected attributes. And no examiner has reviewed any artefact; ``a gate a regulator could accept'' is a design goal, not an obtained acceptance.

\section{The reference run: deciding without evidence}
\label{sec:ref}

CACE-Bench also measures a failure mode that accuracy-style metrics cannot see. Each case requires several classes of evidence, each resolved through a per-jurisdiction provider chain in which a provider may not respond; when the sources needed for the compliance conclusion did not all answer, escalation is the correct outcome. The \emph{silent-decision rate} is the share of such undecidable cases on which the pipeline decided anyway. Table~\ref{tab:ref} reports the reference run (seed~0, 23{,}000 cases, seven Latin American jurisdictions, provider registry 2026-08-03) with the missed-flag rate now reported alongside false positives, and the range over \refnseeds{} seeds.

\begin{table}[t]
\centering\small
\caption{Reference run, CACE-Bench v0.6.0. Blind: availability-blind single pass. Judge: with the judge-and-recovery layer. Seed~0 with 95\% Wilson intervals~\cite{wilson}; denominator $n$; range over seeds 0--\the\numexpr\refnseeds-1\relax.}
\label{tab:ref}
\begin{tabular}{lccccc}
\toprule
Metric (\%) & Blind & Judge & $n$ & Blind, seeds & Judge, seeds\\
\midrule
\input{generated/tab_ref.tex} 
\bottomrule
\end{tabular}
\end{table}

Undecidable cases are \refundec\% of the stream (\refundeclo--\refundechi\% across seeds), and an availability-blind pipeline decides \refsilentoff\% of them. The judge-and-recovery layer brings that to \refsilenton\%. False positives fall from \reffpoff\% to \reffpon\%, and missed flags fall too, from \refmissoff\% to \refmisson\%, so the reduction is not bought by flagging less. The recovery layer, however, is a documented parametric model---each first-pass error is detected with probability 0.85 and corrected with probability 0.92---so this run establishes where the failure concentrates, not how well any particular implementation repairs it.

\section{Threats to validity}
\label{sec:threats}

\textbf{Neither the agent nor the proposer is a language model.} The agent is a deterministic rule that applies the harness, with a 2\% outcome flip standing in for execution error; the proposer is a seeded search over a stated edit space, and the library contains the correct primitive. The track measures the bound and the gate, not proposal quality; an LLM proposer will produce candidates outside this space, including free-text edits whose regression surface is larger. The interface accepts one, and that is the next experiment.

\textbf{Synthetic throughout.} Distributions, shift families and the feedback model (all flags reviewed, 5\% of clears audited) are ours. The no-gate failure depends on misses being less visible than false positives; we believe that asymmetry is realistic, but its size is an assumption.

\textbf{Execution noise is modelled as common random numbers.} Sharing each case's noise draw across harness versions makes paired comparison clean and isolates the gate's response to rule changes. Real LLM noise is not shared across versions, which would raise the noise floor of the regression clause and strengthen the case for a noise-scaled $\epsilon$ (Section~\ref{sec:eps}).

\textbf{The false-admission audit is weak by construction.} $\Hsel$ and $\Hval$ are interleaved halves of the same pre-shift stream labelled by the same deterministic rule, so a rule-level edit that passes on one half will almost always pass on the other. Zero false admissions on $\Hval$ is therefore expected for any gated arm; the informative contrasts are against the no-gate arm and on the post-shift held-out window.

\textbf{Baselines not run.} We did not run a static prompt-optimisation baseline or direct implementations of the cited self-evolution methods; the no-gate arm stands in for the latter's admission logic only.

\textbf{One primitive, one shift at a time.} Interacting primitives and concurrent shifts are not tested.

\textbf{Label source.} The relabelled pool is only as correct as the rule that relabels it.

\textbf{Fairness.} No protected attributes; no fairness claim.

\section{Conclusion}

Self-evolution in regulated lending is acceptable only if what evolves is something a supervisor can review. We argued that the runtime harness is that thing and the only thing, built an engine that respects the bound, and measured the control that enforces it. On a synthetic regulatory-shift track the gate admitted \gadmdual{} of \gpropdual{} candidates with \gfadual{} harmful admissions and recovered every low- and mid-severity cell, while the same loops without it admitted \gadmnogate{} changes, \gfanogate{} of them harmful, and under scope and structural re-interpretations loosened the screen until missed flags exceeded 10\% in \gmtennogate{} of \ncells{} runs. The gate needs labels under the new rule, which makes adaptation supervised at the level of the rule; it divides work between a local and a global loop by the kind of re-interpretation; and its fixed tolerance rejects some large correct changes, which a noise-scaled tolerance should fix. The proposer in these experiments is not a language model, and replacing it with one is the experiment that will show whether the bound survives contact with free text.

\section*{Declarations}

\paragraph{Availability of data and materials.} All code, configuration and per-run outputs, including every gate log, are available in the CACE-Bench repository\ifanon{} (anonymised mirror supplied to the editor)\else{} at \url{https://github.com/rav11l/cace-bench}, release v0.6.0, archived on Zenodo under concept DOI \href{https://doi.org/10.5281/zenodo.21394049}{10.5281/zenodo.21394049}\fi. The data are fully synthetic; no personal or production data were used.

\paragraph{Use of generative AI.} A generative AI assistant (Anthropic Claude) was used in preparing this work: to write and debug parts of the simulation and analysis code, to run the experiments, to draft and edit text, and to check the manuscript against the result files. All research questions, the position argued, the experimental design decisions and the final text are the author's responsibility. Every number in the paper is generated by a script from the released result files rather than typed, and the numbers were independently recomputed from the raw outputs before submission. No generative AI system is a component of the experiments reported: the agent and the proposer in the simulation are deterministic or seeded programs.

\paragraph{Competing interests.} The author declares no competing interests.

\paragraph{Funding.} This research received no external funding.

\paragraph{Authors' contributions.} \ifanon Withheld for review.\else R.A. conceived the study, wrote the software, ran the experiments and wrote the manuscript.\fi

\paragraph{Preprint.} \ifanon A preprint of this manuscript is posted on arXiv; its identifier will be disclosed to the editor.\else This preprint has not been peer reviewed.\fi

\end{document}

%% file: generated/numbers.tex
\newcommand{\missthroracle}{1.9}

\newcommand{\missthrdual}{1.9}

\newcommand{\exscopedegraded}{780}

\newcommand{\missscopeoracle}{1.8}

\newcommand{\missscopenogate}{79.7}
\newcommand{\exscopenogate}{940}

\newcommand{\missscopedual}{1.8}

\newcommand{\fpstructdegraded}{27.4}

\newcommand{\missstructoracle}{1.9}

\newcommand{\fpstructlocalonlycm}{27.4}

\newcommand{\missstructdual}{1.9}

\newcommand{\gpropdual}{7{,}449}
\newcommand{\gadmdual}{144}
\newcommand{\gfadual}{0}

\newcommand{\gmaxadmdual}{4}

\newcommand{\gpropstalegate}{8{,}490}

\newcommand{\gorjstalegate}{470}

\newcommand{\gadmnogate}{550}
\newcommand{\gfanogate}{309}

\newcommand{\gmtennogate}{49}
\newcommand{\gmaxadmnogate}{13}

\newcommand{\famstructlocalonlyprop}{1{,}890}
\newcommand{\famstructlocalonlyreg}{1{,}890}
\newcommand{\famstructlocalonlyadm}{0}

\newcommand{\excstructmiddual}{228}

\newcommand{\excstructmidglobalonly}{156}

\newcommand{\excstructhighdual}{500}
\newcommand{\excstructhighmanual}{408}

\newcommand{\loScopeMidRecovered}{9/10}
\newcommand{\cmScaleMin}{2}
\newcommand{\cmScaleMax}{2}

\newcommand{\staleBothFail}{8{,}490}
\newcommand{\nlogs}{540}
\newcommand{\nlogsok}{540}
\newcommand{\nrecords}{43{,}814}
\newcommand{\hsizemin}{151}
\newcommand{\hsizemax}{253}
\newcommand{\ncells}{90}
\newcommand{\nseeds}{10}

\newcommand{\epsrechighA}{5/10}
\newcommand{\epsfphighA}{18.1}
\newcommand{\epsorjhighA}{24}
\newcommand{\epsrechighB}{10/10}

\newcommand{\wrT}{16099}
\newcommand{\wrDiff}{scope -designation}
\newcommand{\wrImproved}{287}
\newcommand{\wrWorsened}{4}
\newcommand{\wrNtarget}{4{,}561}
\newcommand{\wrGain}{6.2}
\newcommand{\wrGainLo}{5.5}
\newcommand{\wrGainHi}{6.9}
\newcommand{\wrC}{6{,}696}
\newcommand{\wrCreg}{864}
\newcommand{\wrCrate}{12.9}
\newcommand{\wrHash}{0f6978c25079}

\newcommand{\wrSeq}{0}

\newcommand{\wrLocalRej}{21}
\newcommand{\wrGlobalRej}{4}
\newcommand{\wrGlobalRejList}{0.20, 0.40, 0.15, 0.50}
\newcommand{\waT}{17099}

\newcommand{\waDiff}{primitive direct\_stake\_check $\rightarrow$ effective\_stake\_check}
\newcommand{\waGain}{26.3}
\newcommand{\waGainLo}{25.0}
\newcommand{\waCreg}{21}
\newcommand{\waCrate}{0.31}
\newcommand{\waSeq}{25}

\newcommand{\refsilentoff}{55.1}
\newcommand{\refsilenton}{6.5}

\newcommand{\reffpoff}{22.5}
\newcommand{\reffpon}{5.0}

\newcommand{\refmissoff}{15.3}
\newcommand{\refmisson}{3.5}

\newcommand{\refundec}{15.6}
\newcommand{\refundeclo}{15.6}
\newcommand{\refundechi}{16.2}
\newcommand{\refnseeds}{5}

%% file: generated/tab_main.tex
\multicolumn{7}{l}{\textit{Thr shift, mid severity, 10 seeds}}\\
Healthy (no shift) & 2.0 {\scriptsize [1.6, 2.3]} & 1.9 {\scriptsize [1.6, 2.1]} & 0.0 & -- & -- & --\\
Degraded (no adaptation) & 28.0 {\scriptsize [27.3, 28.7]} & 1.9 {\scriptsize [1.6, 2.2]} & 0.0 & 498 & 7.0 & --\\
Oracle update & 1.9 {\scriptsize [1.6, 2.2]} & 1.9 {\scriptsize [1.6, 2.2]} & 0.3 & 0 & 0.0 & --\\
Manual update (lag 2{,}000) & 1.9 {\scriptsize [1.6, 2.2]} & 1.9 {\scriptsize [1.6, 2.2]} & 0.3 & 295 & 4.0 & --\\
No gate (trace check only) & 1.9 {\scriptsize [1.6, 2.2]} & 1.9 {\scriptsize \{1.2--2.5\}} & 0.3 & 139 & 2.5 & 784 / 30 / 6\\
Gate on stale labels & 28.0 {\scriptsize [27.3, 28.7]} & 1.9 {\scriptsize [1.6, 2.2]} & 0.0 & 498 & 7.0 & 948 / 0 / 0\\
Global loop only & 28.0 {\scriptsize [27.3, 28.7]} & 1.9 {\scriptsize [1.6, 2.2]} & 0.0 & 498 & 7.0 & 360 / 0 / 0\\
Local loop only & 1.9 {\scriptsize [1.6, 2.2]} & 1.9 {\scriptsize [1.6, 2.2]} & 0.3 & 81 & 1.3 & 566 / 18 / 0\\
Local only, compute-matched & 1.9 {\scriptsize [1.6, 2.2]} & 1.9 {\scriptsize [1.6, 2.2]} & 0.3 & 74 & 1.0 & 1519 / 20 / 0\\
Dual loop (gated) & 1.9 {\scriptsize [1.6, 2.2]} & 1.9 {\scriptsize [1.6, 2.2]} & 0.3 & 82 & 1.3 & 834 / 21 / 0\\
\midrule
\multicolumn{7}{l}{\textit{Scope shift, mid severity, 10 seeds}}\\
Healthy (no shift) & 2.0 {\scriptsize [1.6, 2.3]} & 1.9 {\scriptsize [1.6, 2.1]} & 0.0 & -- & -- & --\\
Degraded (no adaptation) & 36.9 {\scriptsize [36.1, 37.7]} & 1.8 {\scriptsize [1.6, 2.1]} & 0.0 & 780 & 7.0 & --\\
Oracle update & 2.0 {\scriptsize [1.7, 2.2]} & 1.8 {\scriptsize [1.6, 2.1]} & 0.6 & 0 & 0.0 & --\\
Manual update (lag 2{,}000) & 2.0 {\scriptsize [1.7, 2.2]} & 1.8 {\scriptsize [1.6, 2.1]} & 0.6 & 455 & 4.0 & --\\
No gate (trace check only) & 2.0 {\scriptsize [1.7, 2.2]} & 79.7 {\scriptsize \{2.1--98.7\}} & 38.1 & 940 & 6.9 & 536 / 68 / 50\\
Gate on stale labels & 36.9 {\scriptsize [36.1, 37.7]} & 1.8 {\scriptsize [1.6, 2.1]} & 0.0 & 780 & 7.0 & 921 / 0 / 0\\
Global loop only & 36.9 {\scriptsize [36.1, 37.7]} & 1.8 {\scriptsize [1.6, 2.1]} & 0.0 & 780 & 7.0 & 360 / 0 / 0\\
Local loop only & 3.7 {\scriptsize [0.0, 7.7]} & 1.8 {\scriptsize [1.6, 2.1]} & 0.6 & 266 & 3.0 & 590 / 19 / 0\\
Local only, compute-matched & 2.0 {\scriptsize [1.7, 2.2]} & 1.8 {\scriptsize [1.6, 2.1]} & 0.6 & 111 & 1.0 & 1306 / 20 / 0\\
Dual loop (gated) & 2.0 {\scriptsize [1.7, 2.2]} & 1.8 {\scriptsize [1.6, 2.1]} & 0.6 & 236 & 2.8 & 789 / 20 / 0\\
\midrule
\multicolumn{7}{l}{\textit{Struct shift, mid severity, 10 seeds}}\\
Healthy (no shift) & 2.1 {\scriptsize [1.7, 2.5]} & 1.9 {\scriptsize [1.7, 2.2]} & 0.0 & -- & -- & --\\
Degraded (no adaptation) & 27.4 {\scriptsize [26.8, 28.0]} & 1.9 {\scriptsize [1.6, 2.2]} & 0.0 & 498 & 7.0 & --\\
Oracle update & 2.0 {\scriptsize [1.7, 2.3]} & 1.9 {\scriptsize [1.6, 2.2]} & 0.3 & 0 & 0.0 & --\\
Manual update (lag 2{,}000) & 2.0 {\scriptsize [1.7, 2.3]} & 1.9 {\scriptsize [1.6, 2.2]} & 0.3 & 293 & 4.0 & --\\
No gate (trace check only) & 2.0 {\scriptsize [1.7, 2.3]} & 32.7 {\scriptsize \{1.2--98.1\}} & 16.9 & 1126 & 6.8 & 478 / 88 / 63\\
Gate on stale labels & 27.4 {\scriptsize [26.8, 28.0]} & 1.9 {\scriptsize [1.6, 2.2]} & 0.0 & 498 & 7.0 & 954 / 0 / 0\\
Global loop only & 2.0 {\scriptsize [1.7, 2.3]} & 1.9 {\scriptsize [1.6, 2.2]} & 0.3 & 156 & 2.1 & 43 / 10 / 0\\
Local loop only & 27.4 {\scriptsize [26.8, 28.0]} & 1.9 {\scriptsize [1.6, 2.2]} & 0.0 & 498 & 7.0 & 630 / 0 / 0\\
Local only, compute-matched & 27.4 {\scriptsize [26.8, 28.0]} & 1.9 {\scriptsize [1.6, 2.2]} & 0.0 & 498 & 7.0 & 2133 / 0 / 0\\
Dual loop (gated) & 2.0 {\scriptsize [1.7, 2.3]} & 1.9 {\scriptsize [1.6, 2.2]} & 0.3 & 228 & 3.1 & 878 / 10 / 0\\

%% file: generated/tab_grid.tex
Thr & low & 1{,}600 / 1{,}850 & 13.1 / 1.9 & 2.0 / 1.9 & 13.1 / 1.9 & 2.0 / 1.9 & 2.0 / 1.9 & 2.0 / 1.9\\
Thr & mid & 1{,}940 / 1{,}510 & 28.0 / 1.9 & 1.9 / 1.9 & 28.0 / 1.9 & 1.9 / 1.9 & 1.9 / 1.9 & 1.9 / 1.9\\
Thr & high & 2{,}261 / 1{,}189 & 38.0 / 2.0 & 1.9 / 3.1 & 38.0 / 2.0 & 1.9 / 2.0 & 1.9 / 2.0 & 1.9 / 2.0\\
\midrule
Scope & low & 1{,}815 / 1{,}635 & 23.1 / 1.8 & 2.0 / 69.2 & 23.1 / 1.8 & 2.0 / 1.8 & 2.0 / 1.8 & 2.0 / 1.8\\
Scope & mid & 2{,}223 / 1{,}227 & 36.9 / 1.8 & 2.0 / 79.7 & 36.9 / 1.8 & 3.7 / 1.8 & 2.0 / 1.8 & 2.0 / 1.8\\
Scope & high & 2{,}627 / 823 & 46.3 / 1.9 & 1.9 / 78.8 & 46.3 / 1.9 & 1.9 / 1.9 & 1.9 / 1.9 & 1.9 / 1.9\\
\midrule
Struct & low & 1{,}717 / 1{,}733 & 18.9 / 1.9 & 2.1 / 36.1 & 2.0 / 1.9 & 18.9 / 1.9 & 2.0 / 1.9 & 2.0 / 1.9\\
Struct & mid & 1{,}923 / 1{,}527 & 27.4 / 1.9 & 2.0 / 32.7 & 2.0 / 1.9 & 27.4 / 1.9 & 2.0 / 1.9 & 2.0 / 1.9\\
Struct & high & 2{,}134 / 1{,}316 & 34.5 / 1.8 & 2.1 / 48.0 & 18.1 / 1.8 & 34.5 / 1.8 & 18.1 / 1.8 & 2.1 / 1.8\\

%% file: generated/tab_gate.tex
Dual loop (gated) & 7449 & 144 & 5640 & 1665 & 0 & 25 & 0 & 4\\
Local loop only & 5315 & 111 & 2745 & 2459 & 0 & 2 & 0 & 3\\
Global loop only & 2443 & 25 & 500 & 1918 & 0 & 28 & 0 & 1\\
Gate on stale labels & 8490 & 0 & 8490 & 0 & 0 & 470 & 0 & 0\\
No gate (trace check only) & 5353 & 550 & 4803$^\dagger$ & -- & 309 & 0 & 49 & 13\\

%% file: generated/tab_eps.tex
low & 0.005 & 2.0 {\scriptsize [1.7, 2.3]} & 1.9 {\scriptsize [1.6, 2.2]} & 10/10 & 0 & 0\\
low & 0.010 & 2.0 {\scriptsize [1.7, 2.3]} & 1.9 {\scriptsize [1.6, 2.2]} & 10/10 & 0 & 0\\
mid & 0.005 & 2.0 {\scriptsize [1.7, 2.3]} & 1.9 {\scriptsize [1.6, 2.2]} & 10/10 & 0 & 0\\
mid & 0.010 & 2.0 {\scriptsize [1.7, 2.3]} & 1.9 {\scriptsize [1.6, 2.2]} & 10/10 & 0 & 0\\
high & 0.005 & 18.1 {\scriptsize [6.0, 30.2]} & 1.8 {\scriptsize [1.5, 2.1]} & 5/10 & 24 & 0\\
high & 0.010 & 2.1 {\scriptsize [1.8, 2.3]} & 1.8 {\scriptsize [1.5, 2.1]} & 10/10 & 0 & 0\\

%% file: generated/tab_ref.tex
Silent decision & 55.1 {\scriptsize [53.5{,} 56.7]} & 6.5 {\scriptsize [5.8{,} 7.4]} & 3{,}582 & 54.3--56.1 & 6.5--7.4\\
Compliance FP & 22.5 {\scriptsize [21.8{,} 23.2]} & 5.0 {\scriptsize [4.6{,} 5.3]} & 14{,}948 & 22.2--22.7 & 4.9--5.1\\
Missed flag & 15.3 {\scriptsize [14.3{,} 16.4]} & 3.5 {\scriptsize [3.0{,} 4.0]} & 4{,}470 & 14.6--15.3 & 3.1--3.5\\
Hallucination & 2.6 {\scriptsize [2.4{,} 2.8]} & 0.6 {\scriptsize [0.5{,} 0.7]} & 23{,}000 & 2.5--2.8 & 0.4--0.7\\
Provenance compl. & 93.4 {\scriptsize [93.1{,} 93.7]} & 99.3 {\scriptsize [99.2{,} 99.4]} & 23{,}000 & 93.3--93.8 & 99.3--99.4\\